\documentclass[journal]{IEEEtran}

\usepackage{cite}
\usepackage{graphicx}
\usepackage{url}
\usepackage{amsmath}
\usepackage{amssymb}
\usepackage[caption = false, font=footnotesize]{subfig}
\usepackage{float}
\usepackage{booktabs} 
\usepackage[linkcolor=black]{hyperref}
\usepackage{multirow}
\usepackage{array}
\usepackage{color}
\usepackage{tabularx}
\usepackage{longtable}
\usepackage{tabu}
\usepackage{pifont}
\usepackage{amsmath,epsfig}
\usepackage{algorithm}
\usepackage{algorithmicx}
\usepackage{amsfonts}
\usepackage[noend]{algpseudocode}
\usepackage{threeparttable}
\usepackage{mathtools}
\usepackage{amssymb}
\usepackage{bm}
\usepackage{xcolor}
\usepackage{colortbl} 
\usepackage{makecell}

\newcolumntype{L}[1]{>{\raggedright\let\newline\\\arraybackslash\hspace{0pt}}m{#1}}
\newcolumntype{C}[1]{>{\centering\let\newline\\\arraybackslash\hspace{0pt}}m{#1}}
\newcolumntype{R}[1]{>{\raggedleft\let\newline\\\arraybackslash\hspace{0pt}}m{#1}}
\newcommand{\etal}{\textit{et al}.~}
\newcommand{\ie}{\textit{i}.\textit{e}.}
\newcommand{\eg}{\textit{e}.\textit{g}.}

\makeatother

\ifCLASSINFOpdf

\else

\fi

\newcommand{\bl}[1]{{\color{black}#1}}
\begin{document}
%
\title{2AFC Prompting of Large Multimodal Models \\for Image Quality Assessment}

\author{Hanwei~Zhu*,\thanks{*The authors contributed equally to this work.}
        Xiangjie~Sui*,
        Baoliang~Chen,
        Xuelin~Liu,
        Peilin~Chen, 
        Yuming~Fang,~\textit{Senior Member,~IEEE}, and Shiqi~Wang,~\textit{Senior Member,~IEEE}
\thanks{Hanwei Zhu, Baoliang Chen, Peilin Chen, and Shiqi Wang are with the Department of Computer Science, City University of Hong Kong, Hong Kong, China. (e-mail: \{hanwei.zhu, blchen6-c\}@my.cityu.edu.hk, \{plchen3, shiqwang\}@cityu.edu.hk.)}
\thanks{
Xiangjie Sui, Xuelin Liu, and Yuming Fang are with the School of Information Management, Jiangxi University of Finance and Economics, Nanchang, Jiangxi, China (e-mail: xjsui@foxmail.com, xuelinliu-bill@foxmail.com, fa0001ng@e.ntu.edu.sg).}
\thanks{Corresponding author: \textit{Shiqi Wang}.}

}


\maketitle

\begin{abstract} 
While abundant research has been conducted on improving high-level visual understanding and reasoning capabilities of large multimodal models~(LMMs), their visual quality assessment~(IQA) ability has been relatively under-explored. Here we take initial steps towards this goal by employing the two-alternative forced choice~(2AFC)  prompting, as 2AFC is widely regarded as the most reliable way of collecting human opinions of visual quality. Subsequently, the global quality score of each image estimated by a particular LMM can be efficiently aggregated using the maximum a posterior estimation. Meanwhile, we introduce three evaluation criteria: consistency, accuracy, and correlation, to provide  comprehensive quantifications and deeper insights into the IQA capability of five LMMs. Extensive experiments show that existing LMMs exhibit remarkable IQA ability on coarse-grained quality comparison, but there is room for improvement on fine-grained quality discrimination. The proposed dataset sheds light on the future development of IQA models based on LMMs. The codes will be made publicly available at \url{https://github.com/h4nwei/2AFC-LMMs}.

\end{abstract}


\IEEEpeerreviewmaketitle

\section{Introduction}\label{sec:intro}
\IEEEPARstart{T}{he} \bl{recent breakthroughs in large language models (LLMs) \cite{gpt4} have inspired the development of large multimodal models~(LMMs)~\cite{IDEFICS,ye2023mplug,zhang2023internlm,gpt4v},} aiming to simulate human-like processing of multimodal information. Three representative abilities---instruction tuning~\cite{touvron2023llama}, in-context learning~\cite{brown2020language}, and chain-of-thought prompting~\cite{wei2022chain}---highlight the impressive strides made in this field. With the growing interest in LMMs, especially with the emergence phenomenon of models like GPT-4V~\cite{gpt4v}, it becomes crucial to understand their comprehensive capabilities and limitations.

\begin{figure}[t]
    \centering
    \includegraphics[width=1\linewidth]{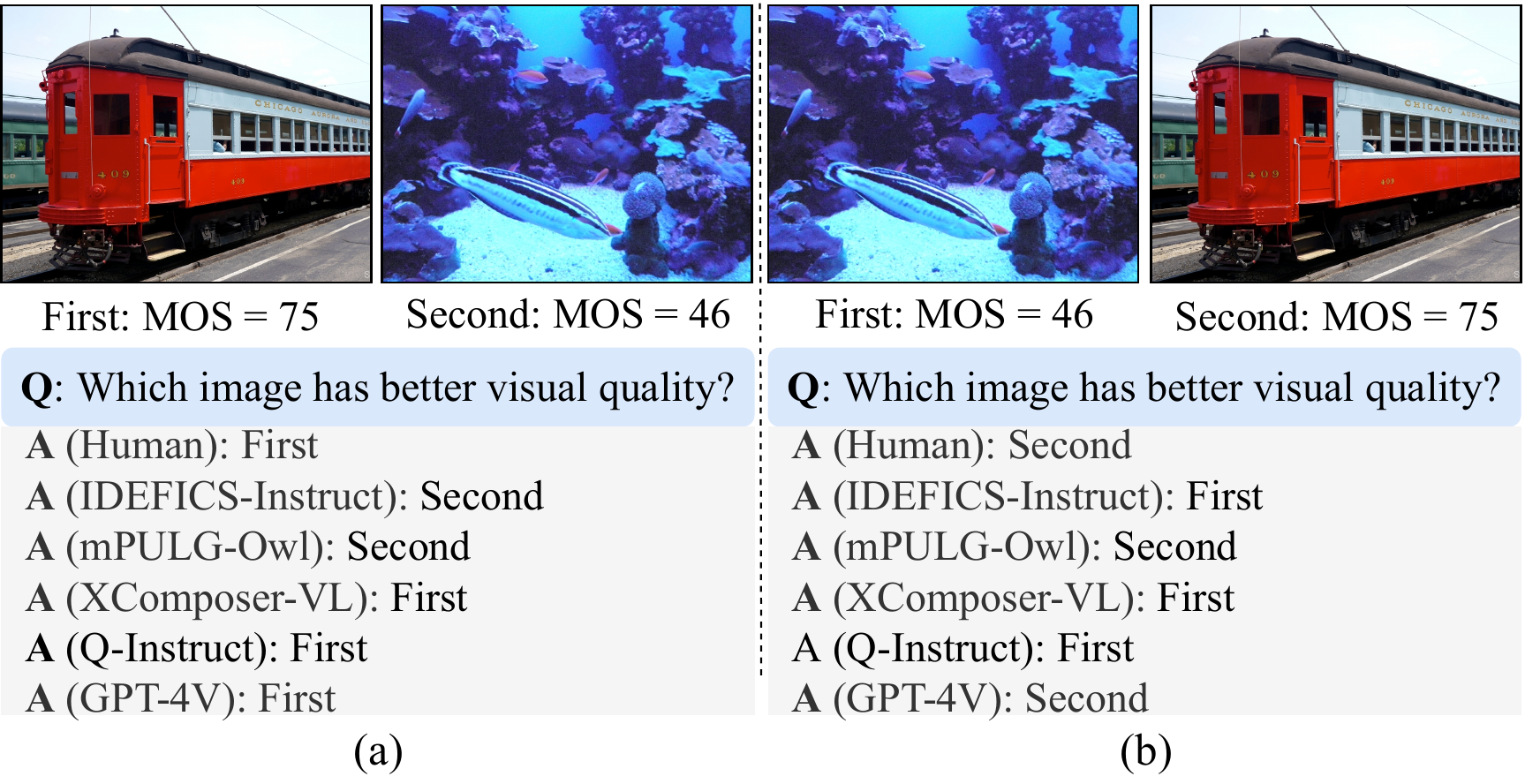}
    \caption{Probing the IQA capability of LMMs via two-alternative forced choice. \textbf{(a)} A pair of images with the corresponding normalized mean opinion scores (MOSs), which is in the range of $[0,100]$. A larger value indicates better visual quality. \textbf{(b)} An order reversed version of (a). Humans can effortlessly select the ``Train'' image with better visual quality regardless of presentation order, but it is unclear whether the LMMs can make the same right choice. In this example, IDEFICS-Instruct~\cite{IDEFICS} gives the incorrect prediction. mPULG-Owl~\cite{ye2023mplug} XComposer-VL~\cite{zhang2023internlm}, and Q-Instruct~\cite{wu2023q} are indifferent to presentation order, and biased towards selecting the second and the first image, respectively. The proprietary  GPT-4V~\cite{gpt4v}  is well aligned with human perception of visual quality.
  }
    \label{fig:motivation}
\end{figure}

The high-level visual understanding and reasoning abilities of LMMs have been extensively evaluated across numerous benchmarks utilizing popular vision-language tasks~\cite{cai2023benchlmm}, such as image captioning, visual question answering, and cross-modality grounding. However, the low-level visual processing and analysis aspects of LMMs remain relatively under-explored. Image quality assessment~(IQA), with the goal of evaluating the perceived quality of visual content, serves as a representative low-level visual task, and holds paramount importance in various image processing~\cite{wu2017blind,yang2023facial}, computer graphics~\cite{zhao2023sptr}, and computer vision applications~\cite{zhu2022learning}. As such, it is highly desirable to evaluate the applicability of LMMs for the IQA task.

\begin{figure*}[t]
    \centering
    \subfloat[AWGN\_Level-$1$]{\includegraphics[width=0.156\linewidth]{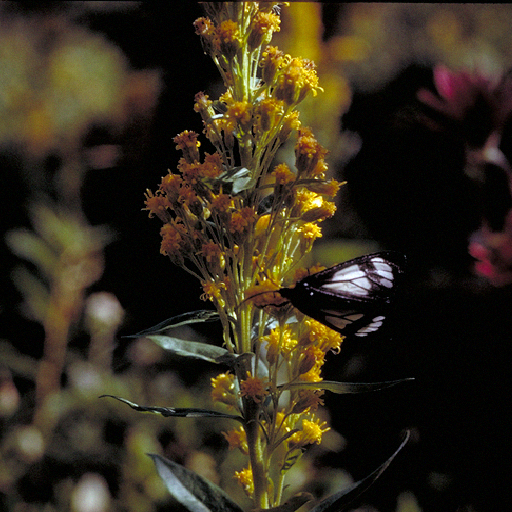}}\hskip.1em
    \subfloat[AWGN\_Level-$2$]{\includegraphics[width=0.156\linewidth]{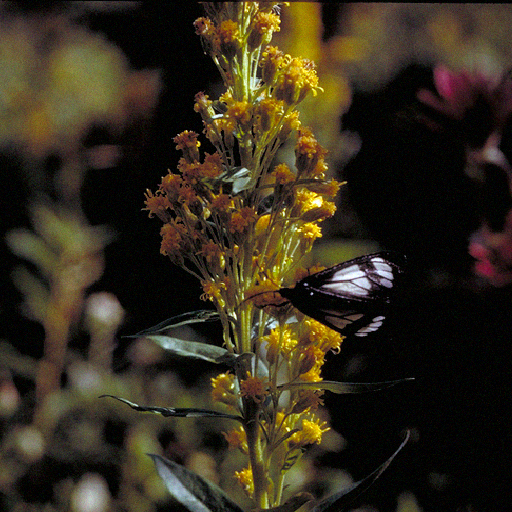}}\hskip.1em
    \subfloat[JP2K\_Level-$4$]{\includegraphics[width=0.156\linewidth]{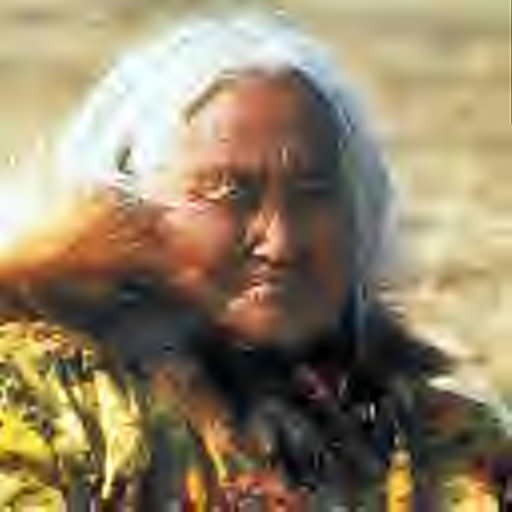}}\hskip.1em
    \subfloat[Pink\_Level-$4$]{\includegraphics[width=0.156\linewidth]{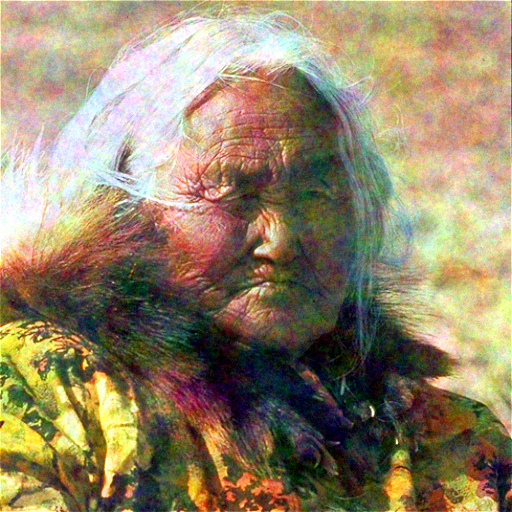}}\hskip.1em
    \subfloat[MOS = $14$]{\includegraphics[width=0.173\linewidth]{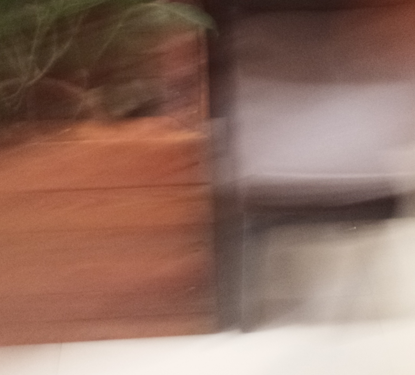}}\hskip.1em
    \subfloat[MOS = $23$]{\includegraphics[width=0.173\linewidth]{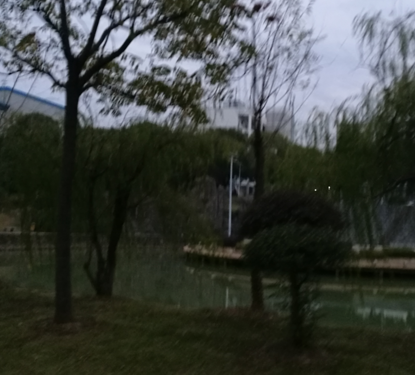}}
    \caption{Illustration of three pairing rules for fine-grained quality comparison. \textbf{(a)}\&\textbf{(b)} Two synthetically distorted images with identical visual content and distortion type but different distortion levels. \textbf{(c)}\&\textbf{(d)} Two synthetically distorted images with identical visual content and distortion level but different distortion types. \textbf{(e)}\&\textbf{(f)} Two realistically distorted images in the MOS interval of $[0, 25)$.  
 }
    \label{fig:fine}
\end{figure*}

Q-Bench~\cite{wu2023qbench} presents an early attempt to assess the visual quality understanding abilities of LMMs through binary quality-relevant question answering, standard quality rating, and quality description. Most recently, You~\etal~\cite{you2023depicting} emphasized on the importance of more human-like quality description over ``contrived'' quality rating, and decomposed IQA into three subtasks: quality description, quality comparison, and comparison reasoning. \bl{Although these studies may seem appealing at first glance, they require human-verified quality descriptions, which is a more costly and time-consuming process than collecting scalar quality ratings. Moreover, the subjective nature of visual quality adds complexity to aggregating descriptions from various users. Last, the semantic comparison of two quality descriptions~\cite{touvron2023llama} represents an unresolved challenge in natural language processing.}

The research in visual quality assessment has a long history, and there are many well-established human-rated image quality databases~\cite{CSIQ2010,KADID2019,MM21,LIVEC2015,koniq10k2020,spaq2020,SQAD2023,KADIS2020,SQAD2023}, readily available to test this perceptual aspect of LMMs. In this paper, we propose to adopt the two-alternative forced choice (2AFC) method, also known as the paired comparison to comprehensively evaluate the IQA capability of LMMs on existing image quality datasets.
We devise coarse-to-fine pairing rules, and use the maximum a posterior~(MAP) estimation~\cite{Tsukida2011} to convert pairwise preferences of different LLMs to the global ranking scores. Additionally, we introduce three evaluation criteria, namely consistency, accuracy, and correlation, to quantify different and complementary aspects of the IQA capability of LMMs. These criteria offer deeper insights into the strengths and weaknesses of LMMs in discriminating image quality variations. Extensive experiments on subsets of eight existing image quality datasets reveal many ``interesting'' behaviors of LMMs (see Fig.~\ref{fig:motivation}). \bl{Most notably, we find that LMMs generally struggle with the IQA task, some of which exhibit strong biases. On the contrary, the proprietary model GPT-4V has exhibited outstanding performance, surpassing the majority of other models by a significant margin. Further testing on more challenging fine-grained pairs reveals that there remains room for improvement.}

\section{Ingredients of Probingg Pipeline}
\label{sec:dataset}

In this section, we first introduce the coarse-to-fine pairing rules, and then detail the maximum a posterior estimation~(MAP)~\cite{Tsukida2011} for multiple options. We last introduce three evaluation criteria to benchmark LMMs.

\subsection{Coarse-to-fine Pairing}
\label{sec:pc}
To evaluate the IQA capability of LMMs, we devise a set of coarse-to-fine pairing rules.
For coarse-grained quality comparison, different images from the same dataset are randomly paired. 
For fine-grained quality comparison, we propose three pairing rules, as illustrated in Fig.~\ref{fig:fine}. The first rule involves pairing synthetically distorted images with identical visual content and distortion type but different distortion levels. The second rule suggests pairing synthetically distorted images with identical visual content and distortion level but different distortion types. The third rule entails pairing realistically distorted images within the same mean opinion score (MOS) interval (\ie, of similar visual quality).

\subsection{Maximum a Posterior Estimation}
For a complete 2AFC design, $N$ test stimuli require $\binom{N}{2}$ paired comparison to derive the global ranking results, which is infeasible when $N$ gets large. As such, we opt for the maximum a posterior estimation~(MAP), which is able to handle $N$ options with fewer needed pairs by solving an optimization problem based on Thurstone's case V model~\cite{Tsukida2011}. We denote a set of images as $\mathcal{D} = \{{x}^{(i)}\}_{i=1}^{N}$ and their MOSs as $\{q^{(i)}\}_{i=1}^{N}$. The LMM is represented by a parametric function $f_\theta$, which takes a textual prompt $t$ and two images $({x}^{(i)}, x^{(j)})$ as inputs, and produces a binary output to indicate whether $x^{(i)}$ is perceived better than $x^{(j)}$. MAP estimation aggregates the quality preference entries $C_{ij}$, which is computed by counting the number of times image $i$ preferred over $j$, into global ranking scores $\{q_\theta^{(i)}\}_{i=1}^{N}$. The log-likelihood of the quality preference matrix $C$ can be defined as 
\begin{align}\label{eq:log}
    \mathcal{L}(q_\theta|C) = \sum_{i,j}C_{i,j}\log(\Phi(q_\theta^{(i)} - q_\theta^{(j)})),
\end{align}
where $\Phi(\cdot)$ is the the standard Normal cumulative distribution function~(CDF). The MAP estimation can be formed by solving the above maximum likelihood function with a ridge regularization on the scale values $p(q_\theta)$:
\begin{equation}
\begin{aligned}
& \underset{q_\theta}{\text{arg max}}
& & \sum_{i,j}C_{i,j}\log(\Phi(q_\theta^{(i)} - q_\theta^{(j)})) - \sum_i\frac{(q_\theta^{(i)})^2}{2} \\
& \text{subject to}
& & \sum_{i}q_\theta^{(i)} = 0.
\end{aligned}
\end{equation}

\begin{figure}[t]
  \centering
  \subfloat{\includegraphics[trim={0.5cm 0cm 1cm 1cm},width=.96\linewidth]{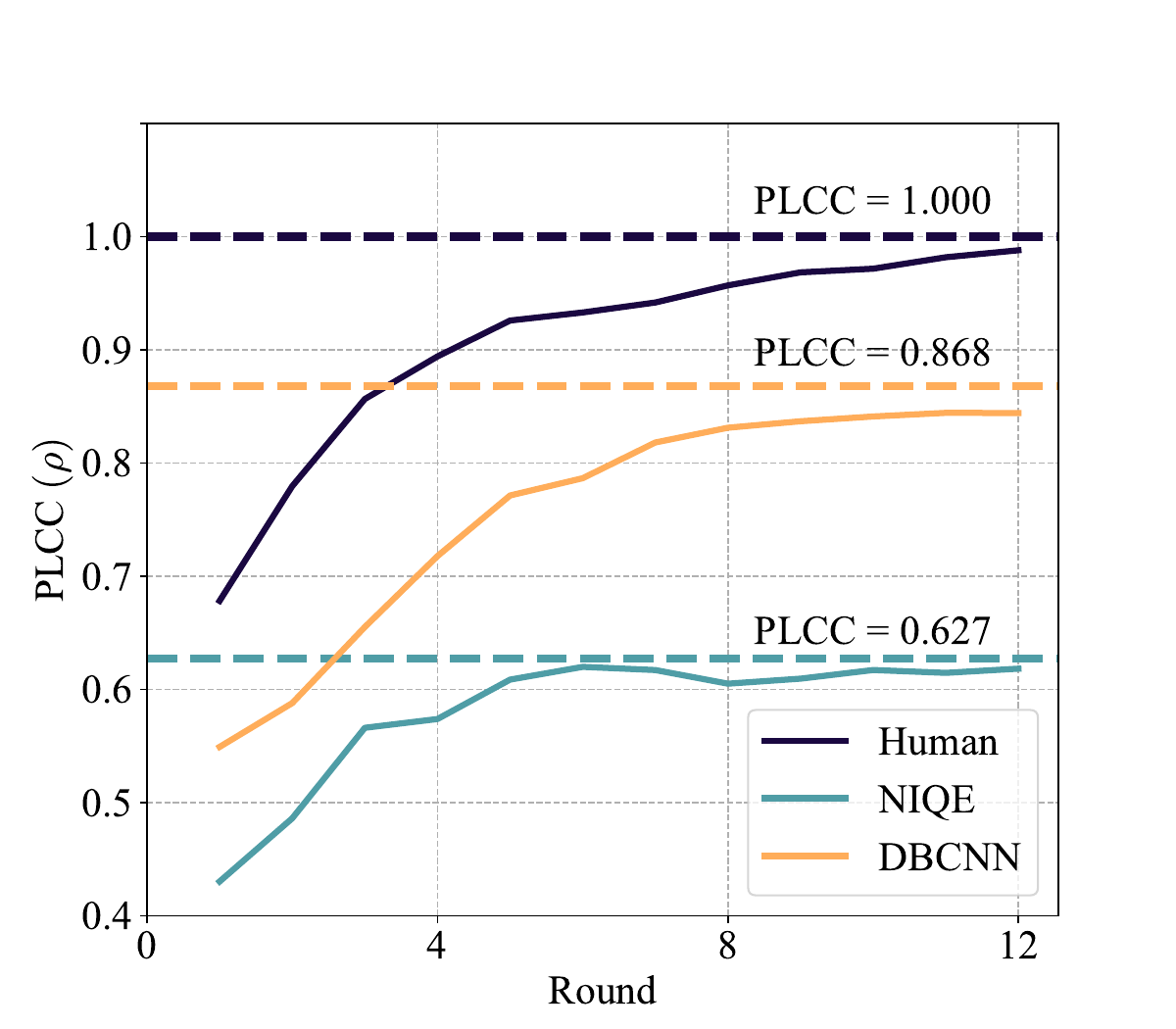}}
  \caption{Validation of MAP estimation in aggregating pairwise rankings from the human observer, NIQE, and DBCNN, respectively. MAP estimation quickly converges as the number of pairing rounds increases, where each round consists of $N$ paired comparisons for $N$ test images. Performance on sampled images from SPAQ ($N = 160$). 
  }
  \label{fig:ts}

\end{figure}

Herein, to verify the reliability and effectiveness of MAP estimation, we adopt MOS (to represent the golden human observer), and two no-reference IQA models NIQE~\cite{NIQE2013} and DBCNN~\cite{DBCNN2020} as the judges to rank image pairs in the SPAQ dataset~\cite{spaq2020}. In particular, each image in sampled SPAQ ($N=160$) is randomly paired with another image in each round, and such pairing process is repeated $M$ rounds until convergence. The winning image shall receive a higher quality score from the judge. We compute the Pearson linear correlation coefficient (PLCC) between the average ranking scores by MAP estimation and MOSs, as shown in Fig.~\ref{fig:ts}. We find that for different judges, MAP estimation quickly converges to the respective upper bounds (represented by the dashed lines), which are calculated using the raw quality scores. Therefore, we conclude that if the outcome of the pairwise comparison is accurate, MAP estimation will produce reliable global quality ranking scores with a manageable number of paired comparison. By employing LMMs as the judges in the MAP estimation, we are able to quantify the capability of LMMs in perceiving visual quality like humans do.

\subsection{Evaluation Criteria}
We propose to quantify the IQA capability of LLMs using the three evaluation criteria.

\noindent\textbf{Consistency}~($\kappa$) measures whether the LLM's prediction is robust to the presentation order of two images:
    \begin{align}\label{eq:consistency}
        \kappa = \frac{1}{\vert\mathcal{P}\vert}\sum_{(x,y)\in\mathcal{P}}\mathbb{I}\left[f_\theta\left( ({x}, {y} ), {t} \right) + f_\theta\left( ( {y}, {x} ), {t} \right)=1\right],
    \end{align}
    where $\vert\mathcal{P}\vert$ denotes the total number of pairs, and $\mathbb{I}[\cdot]$ is the indicator function. $\kappa$ ranges from $[0,1]$ with a larger value indicating higher consistency.

\begin{table}[t]
\caption{Overview of the image subset sampled from eight existing image quality datasets. The two numbers separated by ``/'' in the last two columns represent the number of reference and distorted images, respectively. We emphasize in the third section two datasets---KADIS-700k and SQAD---that are less likely to be included for training LMMs due to the absence of MOSs. We equip them with MOSs from a formal subjective experiment to test LMMs while minimizing the potential data contamination risk
}\label{tab:db}
\centering
\renewcommand\arraystretch{1}
\begin{threeparttable}
\begin{tabular}{c|lccc}
\toprule[0.3mm]
& Dataset & Distortion  & \# images & \# samples  \\ \hline
\multirow{9}{*}{{\makecell[c]{Coarse-\\grained}}} & CSIQ~\cite{CSIQ2010} & Synthetic & $30/866$ & $30$ \\
&KADID-10k~\cite{KADID2019}  & Synthetic & $81/10,125$ & $81$ \\
&MM21~\cite{MM21} & Synthetic & $129/5,031$ & $129$  \\ \cline{2-5}
&CLIVE~\cite{LIVEC2015} & Realistic & $1,169$ & $100$ \\ 
&KonIQ-10k~\cite{koniq10k2020} & Realistic & $10,074$ & $160$ \\ 
&SPAQ~\cite{spaq2020} & Realistic & $11,125$ & $160$ \\ \cline{2-5}
& KADIS-700k~\cite{KADIS2020} & Synthetic & $140$k/$750$k & $100$ \\
& SQAD~\cite{SQAD2023} & Realistic & $3,017$ & $100$ \\ \midrule
\multirow{2}{*}{{\makecell[c]{Fine-\\grained}}} & CSIQ~\cite{CSIQ2010}  & Synthetic & $30/866$ & $4/100$ \\
&SPAQ~\cite{spaq2020} & Realistic & $11,125$ & $100$ \\ 
\bottomrule[0.3mm]
\end{tabular}
\end{threeparttable}
\end{table}

\noindent\textbf{Accuracy}~($\alpha$) measures the accuracy rate of the LLM, provided that the consistency constraint (\ie, $f_\theta\left( ({x}, {y} ), {t} \right) + f_\theta\left( ( {y}, {x} ), {t} \right)=1$) is met:
   \begin{align}
       \alpha = \frac{1}{|\mathcal{P}_s|}\sum_{(x,y)\in\mathcal{P}_s}\mathbb{I}\left[f\left( ( {x}, {y} ), {t} \right) = \mathbb{I}\left[q(x) \ge q(y)\right]\right],
   \end{align}
 where $\mathcal{P}_s\subset\mathcal{P}$ contains the subset of image pairs that the LLM makes consistent prediction. 
   
  \noindent \textbf{Correlation}~($\rho$) measures the linear correlation between the global ranking scores aggregated by MAP and the MOSs:
    \begin{align}
       \rho  = \frac{\sum_{i=1}^{N}(q^{(i)}-\overline{q})(q_\theta^{(i)}-\overline{q}_\theta)}{\sqrt{\sum_{i=1}^{N}(q^{(i)}-\overline {q})^2}\sqrt{\sum_{i=1}^{N}(q_\theta^{(i)}-\overline{q}_\theta)^2}}, 
    \end{align}
    where $\overline{q}$ and $\overline{q}_\theta$ represent the mean MOS and  ranking score, respectively. Before computing $\rho$, a simple monotonic function is commonly applied to map the model prediction to the MOS range as a way of compensating prediction nonlinearity~\cite{video2000final}.

\begin{table*}[htp]
    \caption{VQA capability comparisons of the LMMs in terms of consistency~($\kappa$), accuracy~($\alpha$), and correlation~($\rho$) with the coarse-grained setting.
    The weighted average is computed by the average values weighted by the number of samples of the corresponding subsets. The best two results are highlighted in boldface and underline, respectively}
    \renewcommand\arraystretch{1.2}
    \centering
    \footnotesize
    \tabcolsep=0.12cm
    \begin{threeparttable}
    \begin{tabular}{l|c|ccc|ccc|ccc|ccc|ccc}
    \toprule[0.3mm]
       \multicolumn{2}{c|}{} & \multicolumn{3}{c|}{IDEFICS-Instruct~\cite{IDEFICS}} & \multicolumn{3}{c|}{mPLUG-Owl~\cite{ye2023mplug}} & \multicolumn{3}{c|}{XComposer-VL~\cite{zhang2023internlm}} & \multicolumn{3}{c|}{Q-Instruct~\cite{wu2023q}}  & \multicolumn{3}{c}{GPT-4V~\cite{gpt4v}}\\ \hline
        Dataset & Distortion & $\kappa\uparrow$ & $\alpha\uparrow$ & $\rho\uparrow$ & $\kappa\uparrow$ &  $\alpha\uparrow$  & $\rho\uparrow$  & $\kappa\uparrow$  &  $\alpha\uparrow$ & $\rho\uparrow$ & $\kappa\uparrow$  & $\alpha\uparrow$ & $\rho\uparrow$ & $\kappa\uparrow$  & $\alpha\uparrow$ & $\rho\uparrow$\\  \hline
CSIQ & Synthetic      & $0.206$ & $0.094$ & $0.670$ & $0.422$ & $0.233$ & $0.649$ & $0.233$ & $0.122$ & $0.489$ & $0.117$ & $0.078$  & $0.650$ & $0.778$ & $0.589$ & $0.764$ \\
KADID-10k & Synthetic & $0.202$ & $0.102$ & $0.552$ & $0.396$ & $0.179$ & $0.399$ & $0.267$ & $0.154$ & $0.517$ & $0.387$ & $0.269$ &  $0.466$ & $0.763$ & $0.540$ & $0.560$\\
MM21 & Synthetic      & $0.337$ & $0.173$ & $0.338$ & $0.385$ & $0.204$ & $0.319$ & $0.171$ & $0.109$ & $0.411$ & $0.480$ & $0.324$ & $0.392$  & $0.792$ & $0.544$ & $0.474$\\  \cline{2-17}
CLIVE & Realistic     &$0.323$  & $0.152$ & $0.492$ & $0.365$ & $0.180$ & $0.444$ & $0.133$ & $0.092$ & $0.489$ & $0.327$ & $0.195$ & $0.432$  & $0.837$ &$0.685$  & $0.785$\\
KonIQ-10k & Realistic & $0.251$ & $0.119$ & $0.479$ & $0.399$ & $0.214$ & $0.448$ & $0.148$ & $0.058$ & $0.463$ & $0.489$ & $0.344$ & $0.512$ & $0.836$ & $0.691$ & $0.800$\\ 
SPAQ & Realistic      &$0.330$  & $0.148$ & $0.474$ & $0.332$ & $0.152$ & $0.326$ & $0.208$ & $0.081$ & $0.457$ & $0.485$ & $0.332$ & $0.397$ & $0.871$ & $0.736$ & $0.876$ \\ 
\hline 
\rowcolor{gray!20}
\multicolumn{2}{c|}{{Weighted average}} &  $0.286$ & $0.137$ &  $\underline{0.470}$ & ${0.377}$ &${0.189}$ &  $0.396$ &  $0.188$ &  $0.094$ &  ${0.463}$ & $\underline{0.432}$ & $\underline{0.292}$ & ${0.449}$ &  $\mathbf{0.823}$&  $\mathbf{0.646}$ &  $\mathbf{0.721}$
 \\ \arrayrulecolor{black} \midrule
KADIS-700k & Synthetic & $0.191$& $0.112$  & $0.635$ &${0.388}$&${0.219}$ & $0.612$ & $0.255$ & $0.119$ & $0.596$ & $0.460$ & $0.291$ & $0.658$ & $0.842$ & $0.665$ & $0.674$\\
SQAD & Realistic       & $0.326$& $0.190$  & $0.746$ &${0.330}$&${0.183}$ & $0.618$ & $0.154$ & $0.068$ & $0.690$ & $0.466$ & $0.254$ & $0.642$ & $0.814$ & $0.667$ & $0.742$ \\   \hline
\multicolumn{2}{c|}{Mixed}& $0.296$&$0.133$& ${0.567}$ &${0.372}$& ${0.198}$& $0.503$ & $0.187$ & $0.078$ & $0.500$ & $0.477$ & $0.289$ & $0.503$ &$0.839$  & $0.620$ & $0.678$ \\ 
\hline
\rowcolor{gray!20}
\multicolumn{2}{c|}{{Weighted average}} & $0.289$ & $0.136$	& $\underline{0.629}$ & $ {0.376}$ & ${0.019}$ & $0.559$ & $0.188$ & $0.090$ & $0.572$ & $\underline{0.463}$ & $\underline{0.273}$ & ${0.576}$ & $\mathbf{0.827}$ &  $\mathbf{0.640}$ & $\mathbf{0.693}$\\
     \bottomrule[0.3mm]
    \end{tabular}
    \end{threeparttable}
    \label{tab:fine_performance}
\end{table*}

\section{Experiments}
\label{sec:results}
In this section, we first provide the experimental setups, and then present the coarse-grained and fine-grained IQA results of four LMMs with in-depth analysis. Last, other global ranking aggregation methods are applied to conduct the ablation experiments.

\subsection{Experimental Setups}
We assess five LMMs that accept multiple images as input. These include four open-source models: IDEFICS-Instruct~(based on {LLaMA-9B})~\cite{IDEFICS}, mPLUG-Owl~(based on {LLaMA-7B})~\cite{ye2023mplug}, XComposer-VL~(based on {InternLM-7B})~\cite{zhang2023internlm}, and Q-Instruct~(based on LLaVA\_v1.5-7B)~\cite{wu2023q} and one closed-source model: GPT-4V~\cite{gpt4v}. We initialize the open-source models with their default pretrained weights and use the official API to call GPT-4V.  We sample a total of $1,060$ images from eight image quality datasets to conduct experiments. Synthetically distorted images are selected from CSIQ~\cite{CSIQ2010}, KADID-10k~\cite{KADID2019}, KADIS-700k~\cite{KADIS2020}, and MM21~\cite{MM21} datasets. Realistically distorted images are chosen from CLIVE~\cite{LIVEC2015}, KonIQ-10k~\cite{koniq10k2020}, SPAQ~\cite{spaq2020}, and SQAD~\cite{SQAD2023} datasets. As listed in Table~\ref{tab:db}, the sampled images form two subsets for coarse-grained and fine-grained quality assessment, respectively. It is worth noting that we manually curate $200$ images from KADIS-700k and SQAD datasets, which do not contain MOSs. This selection is to avoid data contamination, considering that closed-source LMM, such as GPT-4V, may have been trained on these image quality datasets. To assign MOSs to images from KADIS-700k and SQAD datasets, we conduct subjective testing using the single stimulus continuous methodology~\cite{bt2002methodology}. We invite $20$ subjects to participate in subjective testing in a controlled laboratory environment, ensuring the ambient illumination does not directly reflect off the displays. We apply the outlier rejection strategy recommended by ITU-T BT.500~\cite{bt2002methodology} to filter noisy data. Finally, the mean value of the valid scores for each image is taken as the ground truth quality score.

For image pairs with inconsistent predictions (see Eq.~\eqref{eq:consistency}), we exclude them from updating the quality preference matrix $C$. It is worth noting that global ranking scores are re-scale to the range of $[0, 100]$. A larger score indicates better visual quality.   We set the round number $M$ to $12$ in our experiments, which is sufficient for the MAP estimation to converge, as shown in  Fig.~\ref{fig:ts}. The pre-defined prompts $t$ are given as follows:\\
\texttt{prompt = [ \\ \textcolor[HTML]{ba2121}{\ `This is the first image:'}, <Image x>,\\
\textcolor[HTML]{ba2121}{\ `This is the second image:'}, <Image y>,\\
\textcolor[HTML]{ba2121}{\ `Which image has better visual quality?'}
\\]}

\subsection{Coarse-grained IQA Performance}
We collect a total of $860$ images for the coarse-grained quality comparison from eight IQA datasets. There are three sampling strategies. 1) We ensure the distorted images selected from the synthetic datasets are content-independent; 2) The mean opinion score~(MOS) distribution of these images of each dataset is maintained as a roughly uniform distribution across five quality levels~(\texttt{excellent}, \texttt{good}, \texttt{fair}, \texttt{poor} \texttt{bad}); 3) We compute the spatial information and colorfulness attributes of the chosen images to ensure the content balance~\cite{winkler2012analysis}.

\begin{table*}
    \centering
    \caption{VQA capability comparisons in terms of consistency~($\kappa$) and accuracy~($\alpha$) with the fine-grained setting}
    \footnotesize
    \begin{threeparttable}
    \begin{tabular}{c|l|cc|cc|cc|cc|cc|cc}
    \toprule[0.3mm]
     \multicolumn{2}{c|}{} & \multicolumn{2}{c|}{IDEFICS-Instruct~\cite{IDEFICS}} & \multicolumn{2}{c|}{mPLUG-Owl~\cite{ye2023mplug}} & \multicolumn{2}{c|}{XCompose-VL~\cite{zhang2023internlm}}  & \multicolumn{2}{c|}{Q-Instruct~\cite{wu2023q}}  & \multicolumn{2}{c|}{GPT-4V~\cite{gpt4v}}  & \multicolumn{2}{c}{Single subject}\\ \hline
        Dataset & Type & $\kappa\uparrow$ & $\alpha\uparrow$ &  $\kappa\uparrow$ & $\alpha\uparrow$ &  $\kappa\uparrow$ & $\alpha\uparrow$ &  $\kappa\uparrow$ & $\alpha\uparrow$ &  $\kappa\uparrow$ & $\alpha\uparrow$ &  $\kappa\uparrow$ & $\alpha\uparrow$ \\  \hline
   
   \multirow{5}{*}{\makecell[c]{CSIQ}} & AWGN & $0.000$ & $0.000$ & $0.448$ &$0.260$&$0.354$&$0.302$&$0.083$&$0.042$&$0.219$&$0.219$&$0.885$&$0.865$\\
                                       & JPEG & $0.000$ & $0.000$ & $0.448$ &$0.250$&$0.448$&$0.177$&$0.073$&$0.073$&$0.490$&$0.459$&$0.885$&$0.865$\\
                                       & JP2K & $0.323$ & $0.323$ & $0.438$ &$0.188$&$0.448$&$0.146$&$0.115$&$0.083$&$0.542$&$0.542$&$0.844$&$0.823$\\ 
                                       & Pink & $0.000$ & $0.000$ & $0.365$ &$0.198$&$0.469$&$0.198$&$0.063$&$0.052$&$0.479$&$0.479$&$0.958$&$0.958$\\
                                       & Blur & $0.156$ & $0.156$ & $0.469$ &$0.167$&$0.479$&$0.146$&$0.240$&$0.156$&$0.364$&$0.313$&$0.979$&$0.958$\\ \hline
   \rowcolor{gray!20}
    \multicolumn{2}{c|}{{Average}}     &  $0.096$ & $0.096$ & $0.433$ &  $\underline{0.212}$ &$\mathbf{0.440}$&$0.194$ &$0.115$&$0.081$&  $\underline{0.419}$ &  $\mathbf{0.402}$&$0.910$&$0.894$ \\\hline
    \multirow{5}{*}{\makecell[c]{CSIQ}} & Level-$1$ & $0.000$ & $0.000$ & $0.552$&$0.292$&$0.177$&$0.073$&$0.104$&$0.042$&$0.010$&$0.000$&$0.406$&$0.354$\\
                                        & Level-$2$ & $0.000$ & $0.000$ & $0.407$&$0.229$&$0.355$&$0.271$&$0.146$&$0.094$&$0.063$&$0.042$&$0.708$&$0.563$\\
                                        & Level-$3$ & $0.000$ & $0.000$ & $0.500$&$0.282$&$0.375$&$0.094$&$0.135$&$0.052$&$0.344$&$0.333$&$0.885$&$0.813$\\ 
                                        & Level-$4$ & $0.125$ & $0.104$ & $0.385$&$0.188$&$0.448$&$0.292$&$0.083$&$0.052$&$0.604$&$0.531$&$0.885$&$0.802$\\
                                        & Level-$5$ & $0.427$ & $0.427$ & $0.458$&$0.219$&$0.458$&$0.208$&$0.115$&$0.104$&$0.604$&$0.313$&$0.958$&$0.927$\\ \hline
      \rowcolor{gray!20}
\multicolumn{2}{c|}{{Average}} &  $0.110$ & $0.106$ & $\mathbf{0.460}$ &  $\underline{0.242}$ &$\underline{0.363}$ &$0.187$ &$0.117$&$0.069$&  $0.325$ &  $\mathbf{0.244}$&$0.769$&$0.692$ \\\hline
   \multirow{4}{*}{\makecell[c]{SPAQ}}  & $[0, 25)$    &$0.465$&$0.174$&$0.417$&$0.201$&$0.153$&$0.104$&$0.389$&$0.236$&$0.465$&$0.208$&$0.896$&$0.472$\\
                                        & $[25, 50)$   &$0.410$&$0.264$&$0.417$&$0.194$&$0.097$&$0.035$&$0.486$&$0.285$&$0.812$&$0.583$&$0.931$&$0.708$\\
                                        & $[50, 75)$   &$0.451$&$0.215$&$0.458$&$0.271$&$0.160$&$0.035$&$0.493$&$0.229$&$0.674$&$0.382$&$0.806$&$0.507$\\ 
                                        & $[75, 100]$  &$0.444$&$0.285$&$0.382$&$0.181$&$0.167$&$0.042$&$0.424$&$0.181$&$0.660$&$0.417$&$0.792$&$0.458$\\ \hline
      \rowcolor{gray!20}
\multicolumn{2}{c|}{{Average}} &  $0.443$ & $0.235$ & ${0.419}$ &  ${0.212}$ & ${0.144}$ &${0.054}$&$\underline{0.448}$&$\underline{0.233}$&  $\mathbf{0.653}$ &  $\mathbf{0.398}$&$0.856$&$0.536$ \\
   \bottomrule[0.3mm]
    \end{tabular}
    \end{threeparttable}
    \label{tab:fine_grain}
\end{table*}

\begin{table}[t]
    \centering
    \caption{VQA capability comparisons in terms of correlation~($\rho$) with the fine-grained setting}
    \tabcolsep=0.14cm
    \footnotesize
    \begin{threeparttable}
    \begin{tabular}{l|ccccc|c}
    \toprule[0.3mm]
 
              & IDEFICS- & mPLUG- &  XCompose- & Q-       &  GPT- & Single  \\  
              & Instruct & Owl    &  VL        & Instruct &  4V   & subject  \\  \hline
    AWGN & $0.000$ & $0.543 $ & $0.586$&$0.449$&$0.815$&$0.919$\\
    JPEG & $0.000$ & $0.482$ & $0.148$ &$0.560$&$0.933$&$0.929$\\
    JP2K & $0.700$ & $0.643$ & $0.189$ &$0.631$&$0.966$&$0.935$\\ 
    Pink & $0.000$ & $0.396$ & $0.085$ &$0.563$&$0.921$&$0.951$\\
    Blur & $0.368$ & $0.396$ & $0.512$ &$0.583$&$0.894$&$0.921$\\ \hline
   \rowcolor{gray!20}
Average  & ${0.214}$ & $0.492$ & $0.304$ &  $\underline{0.557}$ &$\mathbf{0.906}$&$0.931$\\\hline
    Level-$1$ & $0.000$ & $0.444$ & $0.281$&$0.258$&$0.000$&$0.297$\\
    Level-$2$ & $0.000$ & $0.332$ & $0.222$&$0.617$&$0.265$&$0.716$\\
    Level-$3$ & $0.000$ & $0.499$ & $0.409$&$0.664$&$0.492$&$0.769$\\ 
    Level-$4$ & $0.502$ & $0.197$ & $0.375$&$0.230$&$0.855$&$0.910$\\
    Level-$5$ & $0.366$ & $0.216$ & $0.305$&$0.311$&$0.799$&$0.976$\\ \hline
      \rowcolor{gray!20}
Average  &${0.174}$ & $0.338$ & ${0.318}$ &  $\underline{0.416}$ &$\mathbf{0.482}$ &$0.734$\\\hline
    $[0, 25)$  &$0.191$&$0.366$&$0.483$&$0.371$&$0.387$&$0.376$\\
    $[25, 50)$ &$0.198$&$0.373$&$0.326$&$0.503$&$0.412$&$0.646$\\
    $[50, 75)$ &$0.140$&$0.405$&$0.287$&$0.268$&$0.573$&$0.610$\\ 
    $[75, 100]$&$0.395$&$0.372$&$0.346$&$0.169$&$0.420$&$0.346$\\ \hline
      \rowcolor{gray!20}
Average &  $0.231$ & $\underline{0.379}$ & ${0.361}$ &  ${0.328}$ & $\mathbf{0.448}$ &${0.495}$\\
   \bottomrule[0.3mm]
    \end{tabular}
    \end{threeparttable}
    \label{tab:fine_grain_rho}
\end{table}
We conduct the experiments with the images from the same dataset owing to the varying MOS scales across different datasets. The results are shown in~Table~\ref{tab:fine_performance}, from which we can obtain several interesting observations. First, all open-source LMMs, \ie, IDEFICS-Instruct, mPLUG-Owl, XComposer-VL, and Q-Instruct, show poor performance in terms of prediction consistency, suggesting a tendency to provide biased responses regardless of image content. For instance, we count IDEFICS-Instruct and mPLUG-Owl to demonstrate a percentage exceeding $70\%$  of choosing the ``second'' image, while the XComposer-VL exhibits a preference for the ``first'' image with approximately $80\%$. Such biased predictions significantly impact the accuracy and consistency of model inference, leading to inferior performance in IQA. In addition, while Q-Instruct claims to achieve competitive low-level instruction performance for individual images, it shows subpar performance on comparing the visual quality of a pair of images and indicates the model may suffer from over-fitting issues.
Subsequently, GPT-4V outperforms other LMMs across all datasets according to the three proposed evaluation criteria, suggesting its inherent capability to quantify visual quality. To mitigate potential data contamination issues associated with GPT-4V, we strategically select images from the KADIS-700k dataset, which lacks quality-related labels, and the recently released SQAD dataset. Despite these constraints, GPT-4V continues to demonstrate superior performance across these datasets. Notably, the accuracy $\alpha$ of consistent pairs surpasses $60\%$ when compared to human preferences. Moreover, GPT-4V performs better on datasets with realistic distortion compared with synthetic distortion, as shown in Table~\ref{tab:fine_performance}. This could be attributed to the fact that LMMs are trained on the pairs of text and images with realistic distortions. Last but not least, despite the increased complexity presented by pairs that contain mixed distortions, GPT-4V still obtains promising results, outperforming the majority of remaining LMMs.

\begin{table*}[htp]
    \caption{ Quantitative comparison of GPT-4V aggregated by different global ranking methods in terms of PLCC}
    \renewcommand\arraystretch{1.1}
    \centering
    \small
    \tabcolsep=0.3cm
    \begin{threeparttable}
    \begin{tabular}{c|cccccccc}
    \toprule[0.3mm]
    Method & CSIQ & KADID-10k & MM21 & CLIVE & KonIQ-10k & SPAQ & KADIS-700k & SQAD \\
 \hline
    MLE~\cite{Tsukida2011}     & $0.757$ & $0.575$ & $0.472$  & $0.745$ & $0.734$ & $0.828$ & $0.667$ & $0.676$  \\
    Perron~\cite{saaty1984inconsistency}  & $0.708$ & $0.537$ & $0.480$  & $0.760$ & $0.758$ & $0.860$ & $0.650$ & $0.632$  \\
TrueSkill~\cite{Trueskill2007}   & $\mathbf{0.797}$ & $0.557$ & $\mathbf{0.482}$ & $0.773$ & $0.793$ & $\textbf{0.887}$ & $\textbf{0.728}$ & $0.737$  \\\midrule
      MAP~\cite{Tsukida2011}   & $0.764$ & $\mathbf{0.560}$ & $0.474$ & $\mathbf{0.785}$ & $\mathbf{0.800}$ & $0.876$ & $0.674$ & $\mathbf{0.742}$ \\
    \bottomrule[0.3mm]
    \end{tabular}
    \end{threeparttable}
    \label{tab:ablation}
\end{table*}

\subsection{Fine-grained IQA Performance}
We utilize sampled images from CSIQ and SPAQ to carry out the fine-grained quality assessment. Specifically, we sample four scenes from the CSIQ dataset. Each of them is degraded by five classical distortion types, including additive white Gaussian noise (AWGN), JPEG, JPEG 2000~(JP2K), pink noise, and blur, and five distortion levels. For realistic distortion, we select SPAQ as our target database, partition the normalized MOS scores at the range of $[0, 100]$ into four equidistant intervals, and strictly limit the sampling to $25$ images per MOS interval.

Since fine-grained quality comparisons are more challenging, we invite one subject to conduct the experiment in the same setting as the four LMMs. We present the consistency and accuracy in Table~\ref{tab:fine_grain}. The correlation results are shown in Table~\ref{tab:fine_grain_rho}.
For the synthetically distorted images from the CSIQ dataset, we first compare the images with the same content and identical distortion types but varied distortion levels. From Tables~\ref{tab:fine_grain},~\ref{tab:fine_grain_rho} and Fig.~\ref{fig:fine}~(a)\&(b), we observe that the existing LMMs struggle to distinguish the image with superior quality in such fine-grained comparisons. Although GPT-4V outperforms other LMMs, the average consistency and accuracy values are far from human performance. The AWGN is the most difficult distortion type compared with other distortions for the existing LMMs, possibly due to its rarity in the training data. The correlation of Q-Instruct underperforms IDEFICS-Instruct, indicating that the low-level information of single images does not improve the visual quality comparison capability of the Q-Instruct. Subsequently, we compare the images with the same content and distortion levels but different distortion types. We find that it is challenging to compare a pair of images with slight distortions (\eg, Level-1 and Level-2), and the LMMs recognize them as the same image. However, as the level of distortion increases, different distortions demonstrate distinct visual appearances, making it easier to discern the superior quality image (see Fig.~\ref{fig:fine}~(c)\&(d)). Additionally, mPLUG-Owl outperforms GPT-4V in terms of the consistency criterion and achieves the second-best average accuracy value, which reveals that mPLUG-Owl demonstrates better fine-grained IQA capability compared with other open-source LMMs. For realistically distorted images from the SPAQ dataset, we present four normalized MOS intervals to conduct the fine-grained comparison. Most LMMs perform better for realistic distortions than synthetic distortions. Subjects and LMMs can more readily discern preferences within the quality interval of $[25, 50)$. It is reasonable for LMMs to exhibit inferior performance in the quality interval of $[0,25)$ and $[75, 100]$ as they represent either severe distortion or high-quality visual appearance, both of which pose significant challenges for human observers in selecting an image of superior quality. Overall, GPT-4V demonstrates the best performance in the fine-grained IQA experiment, but there remains ample room for improvement.

\subsection{Ablation Experiments}
In this subsection, we compare the MAP estimation to other global aggregation methods, including the maximum likelihood estimation~(MLE)~\cite{Tsukida2011}, the Perron rank method~\cite{saaty1984inconsistency}, and the TrueSkill rating system~\cite{Trueskill2007}. In contrast to MAP estimation, MLE directly optimizes the log-likelihood function to derive the global ranking scores. The Perron rank method uses the principal eigenvector of a pairwise comparison matrix to generate the global ranking, providing a robust and efficient solution for ranking problems~\cite{saaty1984inconsistency}. TrueSkill rating system is a video game ranking algorithm that models the skill of each player as a univariate Gaussian random variable, with mean and variance representing the average skill of each player and the degree of uncertainty, respectively. As shown in Table~\ref{tab:ablation}, we can observe MAP and TrueSkill perform better than other methods. While TrueSkill demonstrates competitive results to MAP, we continue to choose MAP estimation as our default method owing to its mathematically appealing properties.

\section{Conclusions}

In this work, we have probed the low-level IQA capabilities of LMMs, an area that has been relatively under-explored. We devise coarse-grained and fine-grained pairing rules to evaluate the IQA ability of the state-of-the-art LMMs. The MAP estimation is used for aggregating global ranking scores of different LMMs. We further propose three evaluation criteria, providing a comprehensive evaluation of the LMM's IQA abilities. Our analysis has revealed that most LMMs generally lack IQA capabilities and tend to exhibit biased preferences. The proprietary LMM, GPT-4V, shows promising performance in the coarse-grained subset, indicating potential applicability in IQA. However, there is still considerable room for improvement in the IQA ability of existing LMMs, particularly in the fine-grained subset. We hope that our benchmark and analysis will serve as a catalyst for the development of more advanced and versatile LMMs in the field of visual quality assessment.

\section*{Acknowledgements} The authors thank Kede Ma for his insightful discussions and diligent efforts in revising the paper.

\ifCLASSOPTIONcaptionsoff
  \newpage
\fi

\bibliographystyle{IEEEtran}
\bibliography{trans}

\end{document}